\documentclass[a4paper, 10pt, conference]{ieeeconf}      

\IEEEoverridecommandlockouts                              

\usepackage{graphics} 
\usepackage{epsfig} 
\usepackage[subrefformat=parens]{subcaption}
\usepackage{mhchem}
\usepackage{here}
\usepackage[style=ieee]{biblatex}
\usepackage{color}

\title{\bf \LARGE 
Safe mobility support system using crowd mapping\\
and avoidance route planning using VLM
}

\author{Sena Saito$^{1}$, Kenta Tabata, Renato Miyagusuku and Koichi Ozaki
\thanks{This research was supported by JST Program for co-creating startup ecosystem, Grant Number JPMJSF2319, Japan.
This work was also supported by the Graduate Student Research Encouragement Grant of Utsunomiya University (2024).}
\thanks{$^{1}$Sena Saito, Kenta Tabata, Renato Miyagusku and Koichi Ozaki are with Graduate School of Regional Development and Creativity, Division of Engineering and Agriculture,
Graduate School, Utsunomiya University, Tochigi, Japan}
        {\tt\small s.saito@irlab-uu.jp}%
}

\begin{document}

\maketitle
\thispagestyle{empty}
\pagestyle{empty}

\begin{abstract}
Autonomous mobile robots offer promising solutions for labor shortages and increased operational efficiency. However, navigating safely and effectively in dynamic environments, particularly crowded areas, remains challenging. This paper proposes a novel framework that integrates Vision-Language Models (VLM) and Gaussian Process Regression (GPR) to generate dynamic crowd-density maps (``Abstraction Maps'') for autonomous robot navigation.  Our approach utilizes VLM's capability to recognize abstract environmental concepts, such as crowd densities, and represents them probabilistically via GPR. Experimental results from real-world trials on a university campus demonstrated that robots successfully generated routes avoiding both static obstacles and dynamic crowds, enhancing navigation safety and adaptability. 
\end{abstract}

\section{Introduction}
In recent years, the progression of an aging society and declining birthrates have become serious social issues leading to labor shortages. With the growing demand for labor alternatives and the development of robotic technologies, autonomous mobile robots are expected to perform a wide range of tasks efficiently in industries such as logistics and services\cite{alatise}. However, for these robots to operate safely and efficiently in real-world environments, it is essential to have dynamic, real-time environmental perception and flexible path planning. However, recognizing abstract concepts that change moment by moment—such as crowds—is a significant challenge. Vision-Language Models (VLMs) are capable of processing both visual and textual information in an integrated manner and can recognize abstract concepts (like crowds) that have been difficult to judge with conventional methods. Detecting and analyzing such abstract concepts enables the assignment of diverse tasks to robots.

For robots to perform tasks efficiently, it is necessary to obtain the most efficient path to the destination. In real-world scenarios, however, it is not enough to simply search for the shortest route; various types of environmental information must be considered to determine an appropriate travel path. For instance, even when two routes share an equivalent distance, opting for the one characterized by heavier traffic may result in longer travel times. This factor is crucial for ensuring that the robot can reliably accomplish its tasks. One representative algorithm for route searching is Dijkstra's algorithm \cite{dijkstra}. Setting the travel costs, this algorithm can find the optimal path. Typically, autonomous mobile robots use a two-dimensional grid map and calculate a simple shortest path. In such cost calculations, geometric information such as static and dynamic obstacles is mainly used, meaning that abstract environmental concepts like road surface conditions and crowd presence are not represented. In this paper, we propose a method that creates a crowd map using a VLM, combines geometric information with abstract environmental data to generate a cost map, and enables the robot to pre-plan a crowd-avoidance path that results in a safe travel route.

\begin{figure*}
    \centering
    \includegraphics[width=0.9\linewidth]{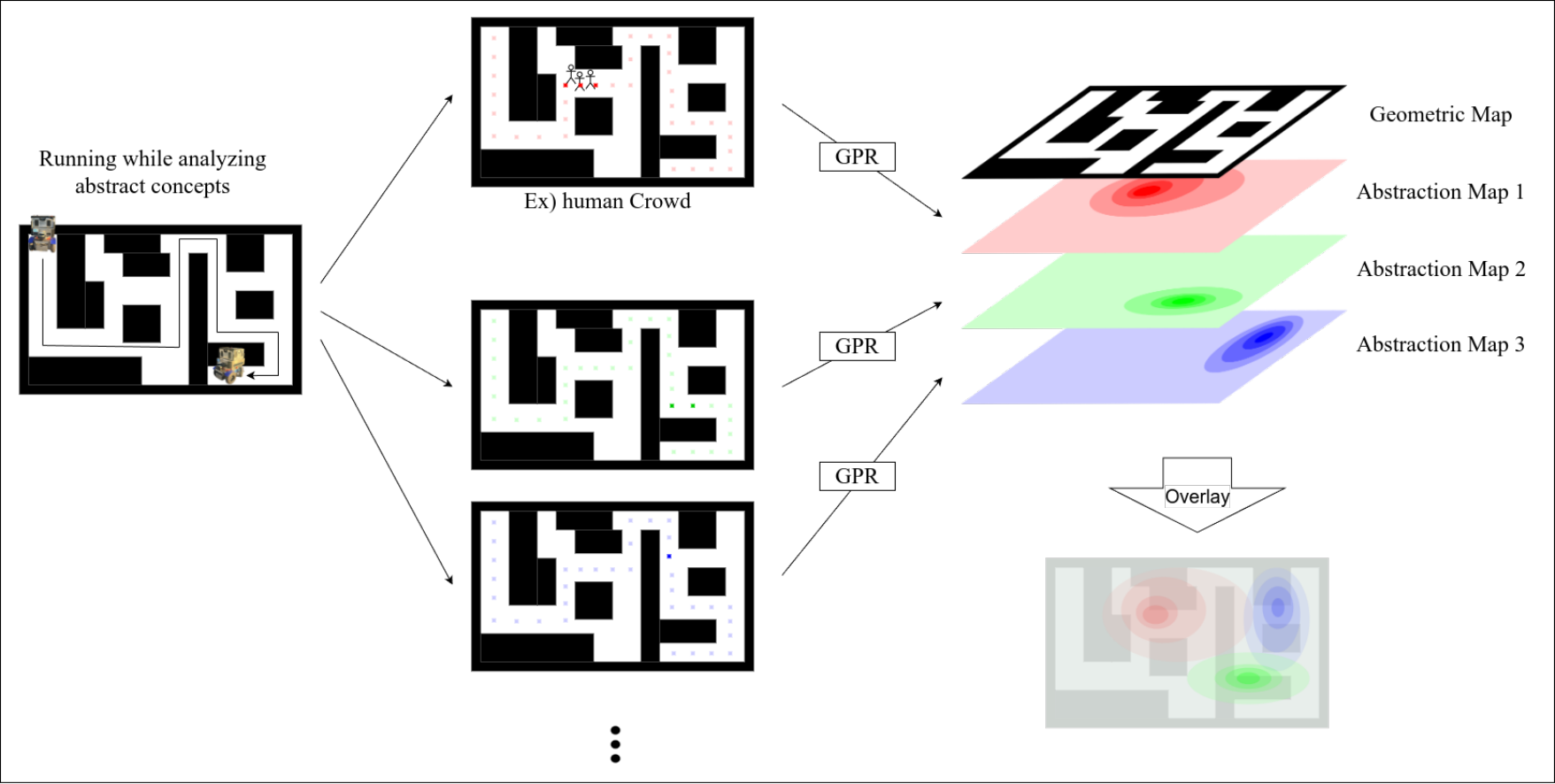}
    \caption{Proposed our concept of Abstraction Map Generator(AMG)}
    \label{fig:method_concept}
\end{figure*}

\section{Related Works}
\label{related}
Congestion analysis is recognized as an important research topic in applications such as smart cities and smart buildings. In particular, numerous studies have focused on methods for counting people within crowds. Approaches for crowd counting include processing images captured by cameras using AI, estimating crowd counts and positions based on Wi-Fi signal strength and reception conditions, and even estimating crowds from environmental data such as temperature and humidity. Each of these methods faces various challenges including limited scope of application and high implementation costs \cite{depatla}.

\subsection{Congestion Analysis Using Various Sensors}
\label{related-sensor}
Choi et al. proposed a method that estimates both the number and locations of people in a crowd using Wi-Fi sensing \cite{choi}. Their approach utilizes the Channel State Information (CSI) from a Wi-Fi router (ESP32) and applies reinforcement learning for estimation. Jiang et al. proposed an indoor occupancy estimation method based on carbon dioxide measurements \cite{jiang}. They designed a learning model called FS-ELM, which collects \ce{CO2} concentration and past occupancy states at fixed intervals to estimate the current number of occupants. Because ELM operates very quickly, it is capable of real-time performance. Devices such as Wi-Fi sensors and \ce{CO2} detectors can be used even in low-light or occluded environments without concerns over privacy, unlike vision-based methods. However, applying these systems to robots requires installing special devices either on the robot itself or within the robot’s operating environment. In addition to analyzing crowds, cameras can be used for multiple tasks such as obstacle avoidance and collecting other environmental information. For this reason, camera have been installed as standard feature for autonomous mobile robots, making them a good options for robots to recognize crowds recognition.

There are several examples of crowd counting and recognition using cameras. For instance, Shami et al. proposed a head-detection-based crowd estimation method for extremely dense crowds \cite{shami}. In their approach, the input image is divided into elliptical patches, and the number of people is estimated for each patch based on head size, avoiding direct regression. Ma et al. introduced Bayesian Loss into a standard CNN and proposed a weakly supervised crowd estimation method based on expected counts \cite{ma}. Instead of performing pixel-level density map regression, their method uses the expected count at each annotation point, making it robust to occlusion and perspective changes. Sharma et al. incorporated a self-attention module into a multi-scale crowd density estimation method \cite{sharma}. Their dual-branch structure extracts features from both the original-resolution and downsampled images and then integrates the correlations between these features, allowing the approach to adapt to different scales. Thanks to advances in algorithm development and high-performance computing, these vision-based methods have achieved increasing accuracy and have demonstrated excellent performance on various datasets. Even in problems that demand high-precision crowd counting, these methods can outperform human capabilities. However, many of these datasets are based on overhead images captured by monitoring cameras or drones, and it is rare for typical autonomous mobile robots to utilize such overhead images for task planning.

Gong et al. applied three different methods (Face++, Darknet YOLO, and Cascaded) to both overhead images and front images (e.g., selfie images) and analyzed their characteristics \cite{gong}. Their results indicated that CNN-based methods can more accurately capture crowds from overhead images than from front images. In overhead images, the overall distribution of the crowd can be grasped at a glance, making it easier to quantify the level of congestion; however, in front images, the overlapping of faces and bodies, especially in dense settings, makes it difficult to identify individual people. For robotic navigation, precise crowd density is not always necessary—a robot only needs to judge whether an upcoming route is passable, so the detailed counting and density estimation does not required. Moreover, in computer vision research, many datasets are processed with extremely large models that improve estimation accuracy but are specifically tuned for crowd analysis and may not extend to other applications.

\subsection{Crowd-Based Navigation in Robotic Systems}
\label{related-navi}
In the context of autonomous mobile robots, navigating through crowds is known to be extremely challenging. Numerous studies have addressed path planning for avoiding people within densely populated environments. Trautman et al. proposed a trajectory modeling method based on Interacting Gaussian Processes (IGP) to enable robots to navigate safely and efficiently in dense crowds \cite{trautman1}\cite{trautman2}. Instead of predicting each agent (robots or people) independently, they modeled them as continuous stochastic processes, thus capturing interdependencies among agents simultaneously. Chen et al. highlighted the high computational cost when applying model-based probabilistic methods in high-density environments and aggregated interactions into a compact crowd model using a self-attention mechanism \cite{changan}. Vemula et al. built a model that learned natural cooperative behavior in crowds using the ETH pedestrian video dataset \cite{anirudh}. Liu et al. refined the optimal action policy using deep reinforcement learning after training on human driving demonstrations \cite{liu}. These methods treat the crowd as a probabilistic model or network and address the problem of navigating through it, which requires predicting either individual human behaviors or the collective behavior of the crowd—necessitating complex computations. Most approaches in autonomous navigation have focused solely on avoiding collisions with people.

Approaches to handling crowds are not limited merely to avoidance; some research also promotes coexistence between robots and humans. Henry et al. used inverse reinforcement learning to estimate overall human flow from observed walking trajectories, and developed a planner that enabled a robot to either follow the flow or move against it as necessary \cite{henry}. Truong et al. integrated the Extended Social Force Model (ESFM) with the Hybrid Reciprocal Velocity Obstascle (HRVO) to enhance a human comfort index in robot navigation \cite{truong}. Nonetheless, these studies have generally assumed that the robot must moves through the crowd and have not fundamentally considered strategies for avoiding crowded areas altogether. This is because it is challenging to quantitatively define a ``crowd''. For a robot, a crowd is not determined solely by the number of people but also by complex factors such as the width of the passage. Conventional computer vision techniques have struggled to analyze such abstract concepts and incorporate that information into autonomous navigation.

The literature we reviewed presents a variety of methods for crowd analysis and crowd-based navigation. However, practical approaches for enabling robots to accurately understand the abstract concept of crowds and effectively incorporate that information into navigation have yet to be fully established. In this study, we therefore investigate a new method for determining the presence of crowds using a VLM and integrating that information into navigation.

\begin{figure}[b]
    \centering
    \begin{minipage}{0.7\linewidth}
        \centering
        \includegraphics[width=\linewidth]{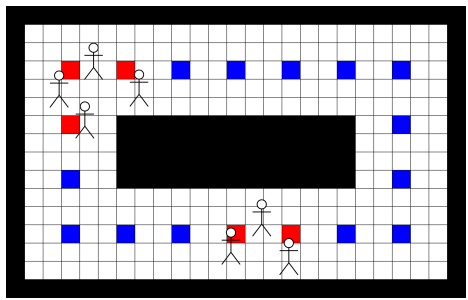}
        \subcaption{Binary ocupancy of obstacle and crowd on the grid map}
        \label{fig:GPR_1}
    \end{minipage}
    \begin{minipage}{0.7\linewidth}
        \centering
        \includegraphics[width=\linewidth]{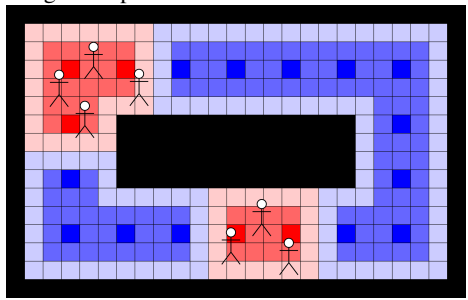}
        \subcaption{Smoothed gridmap by GPR}
        \label{fig:GPR_2}
    \end{minipage}
    \caption{Caliculation algorithm for path planning}
    \label{fig:GPR}
\end{figure}

\section{Abstraction map generation(AMG) method and path planning strategy with the map}
Our proposed method is shown in Fig. \ref{fig:method_concept}. In our approach, images captured by a camera mounted on the robot and a prompt are input into a VLM. The VLM outputs a binary indicator representing the presence or absence of abstract information (in this experiment, indicating the presence or absence of a crowd) at the robot's current location. The robot performs this mapping every 3 meters, representing the crowd cost map on a grid. Then, interpolation is carried out between the measured cost values using Gaussian Process Regression (GPR)\cite{seeger}. GPR generates a distribution with a probability density function that is then reflected in the map. Finally, the Abstraction Map and the geometric map are merged, and the cost information is used with Dijkstra's algorithm to plan a route. Representing both the geometric and abstraction maps with grid map, as shown in Fig. \ref{fig:GPR}, it leads to seamless integration for path planning is achieved.

\subsection{Confidence Evaluation of the Abstraction Map Using Gaussian Process Regression}
GPR was selected to estimate the function $y=f(x)$ that maps an input variable $x$ to a real number $y$ since GPR is a probabilistic model based on Gaussian distributions, its output provides not only the mean prediction for the input data but also the prediction's variance\cite{akaho}. A smaller variance indicates lower uncertainty and thus higher confidence in the prediction. By applying GPR within the Abstraction Map Generator, the variance of the probability density function at a given point provides a measure of confidence in that information. \textcolor{black}{In addition, the proposed method employs a Gaussian kernel (RBF kernel) as the kernel function. Crowd distributions are expected to vary relatively smoothly in both time and space, and the Gaussian kernel is suitable in that it is infinitely differentiable and provides a strong guarantee of smoothness.} This enables integrated decision-making in robot navigation based on the confidence levels of the abstraction information at each location.


\subsection{Environmental Maps}
Although a robot can plan the shortest path to its destination based solely on geometric information, taking the shortest route is not always best route. For example, crowds being dynamic obstacles not represented in the geometric map are fluid and their evaluation is complex. Moreover, in real-world environments where autonomous mobile robots are expected to operate, such situations frequently occur. By also considering factors such as road conditions and the state of other traffic participants, more intelligent path planning can be achieved. When comprehensive environmental information is expressed as cost values, its influence can be reflected in the planned route. As shown on the right side of Fig. \ref{fig:method_concept}, adopting a multi-layer structure for the Abstraction Map allows the preservation of multiple independent types of information, and users can determine which information to prioritize by adjusting the weights during fusion. In this study, to validate the basic characteristics of Abstraction Map Generator (AMG), only the crowd map is used as the source of environmental information for robot path planning.

\section{Experiments on Crowd-Avoidance Navigation Using AMG}

\subsection{Experimental Conditions}
\begin{figure}[t]
    \centering
    \includegraphics[width=0.8\linewidth]{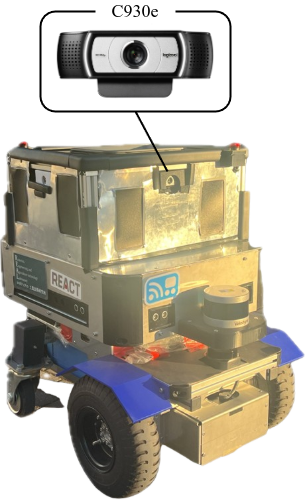}
    \caption{Robot used in the experiment}
    \label{fig:kosobot}
\end{figure}

\begin{figure}[t]
    \centering
    \includegraphics[width=0.6\linewidth]{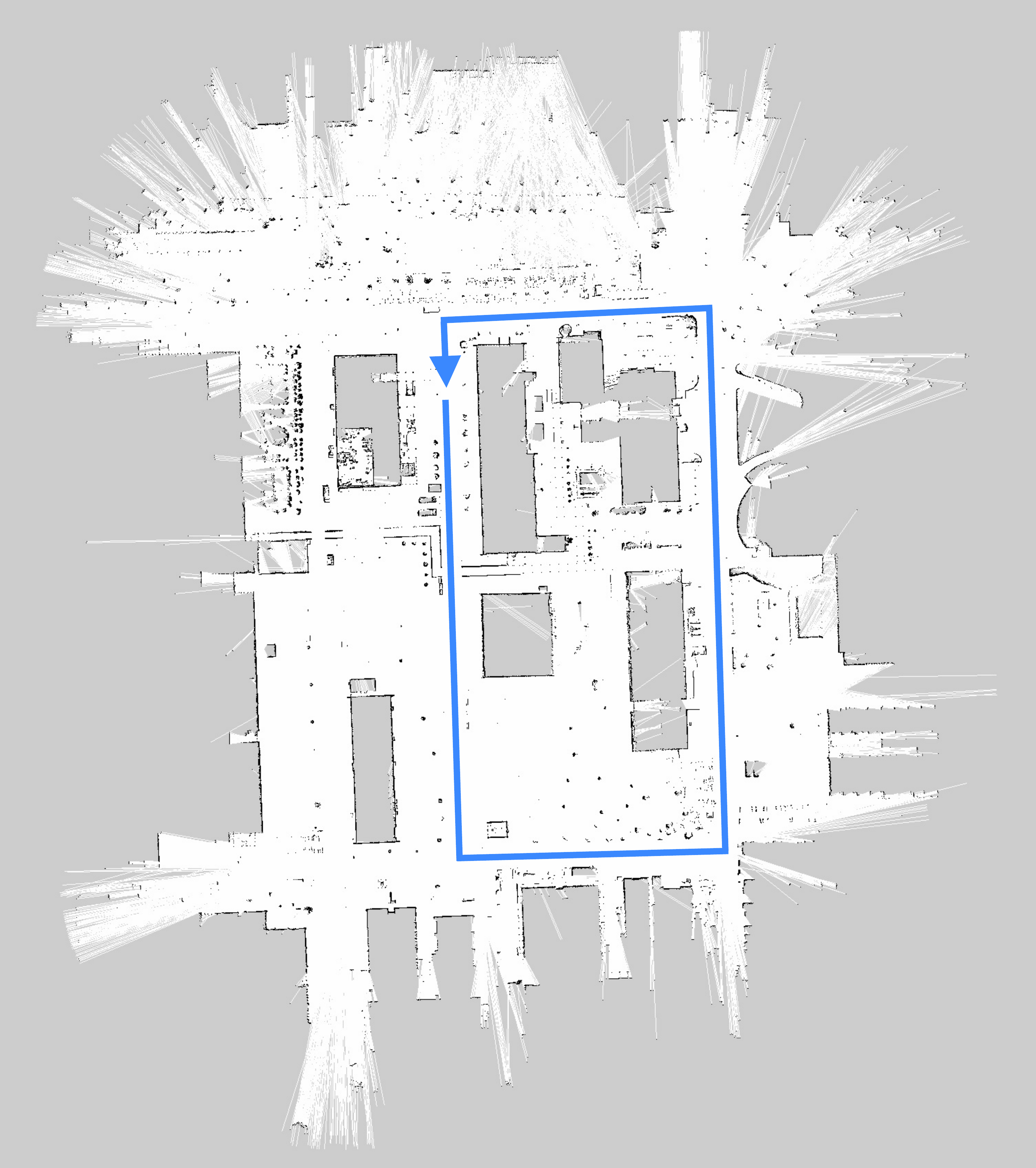}
    \caption{Geometric map in the experiment}
    \label{fig:campus_map}
\end{figure}

To evaluate the proposed method, experiments of crowded map generation were conducted. We make intentionally local crowd on the university campus to evaluate the effect of environmental map creation and its influence on path planning. The experiments used the robot shown in Fig. \ref{fig:kosobot}. This robot is equipped with a 3D-LiDAR for map creation, self-localization, and path planning and RGB camera mounted at the front. Other specifications are shown in TABLE \ref{table:kosobot}. The experiments field is the Utsunomiya University campus. We chose a time when there were few people and intentionally created a crowd so that its effects were noticeable on the crowded map. Fig. \ref{fig:campus_map} shows the geometric map of the campus created by the robot. In the first trial, a crowd was intentionally generated near location \textcircled{\scriptsize 1}, and in the second trial, near location \textcircled{\scriptsize 2}. Based on the robot's self-localization results and the image data, a crowd map was created and an crowded map was generated. In this experiment, gpt-4o-mini was used as the VLM for AMG. \textcolor{black}{This model can handle both language and vision as input simultaneously, allowing for flexible design.}

\begin{table}[t]
    \centering
    \caption{Robot Specifications}
    \label{table:kosobot}
    \begin{tabular} {c|c}
        \hline
        Hardware                & Model Number \\
        \hline
        Size [m]                & \begin{tabular}{c}
                                    Width = 0.58 \\
                                    Depth = 0.72 \\
                                    Height = 0.67 \\
                                  \end{tabular}\\
        Weight [kg]             &   84.4 \\
        PC                      & \begin{tabular}{l}
                                    ASUS PN50 \\
                                  \end{tabular}\\
        IMU                     & ICM-20948(TDK InvenSense) \\
        3D-LiDAR                & VLP-16 (Velodyne Lidar) \\
        2D-LiDAR                & RPLIDAR  S2M1 ×2 (Slamtec) \\
        Motor                   & BLMR5100K-50FR (Oriental motor) \\
        Motor Driver/Encorder   & BLVD-KRD (Oriental motor) \\
        Battery                 & IFM24-400E2 (O'CELL) \\
        RGB Camela              & Logicool C930e \\
        \hline
    \end{tabular}
\end{table}

\subsection{Results of Map Generation Using AMG and the Resulting Planned Paths}
The analysis results using the prompt are shown in TABLE \ref{tab:result}.

\begin{itemize}
    \item Case 1: The crowd is captured from a distance.
    \item Case 2: The robot approaches the crowd, crowd clearly obstructs the path.
    \item Case 3: The robot has passed through the crowd, and no other obstacle obstructs the upcoming path.
\end{itemize}

The images shown are excerpts from the first trial, representing the images input to the VLM in each case. The same fixed prompt was provided in all cases, VLM returns ``crowd'' if a crowd is detected and ``free'' otherwise. \textcolor{black}{In this experiment, we focused on whether intelligent navigation can be realized by combining crowd information and geometric information. Therefore, the prompt design was kept extremely simple, but the prerequisites for scene understanding were clarified by adding the shooting viewpoint information of the image.} In the crowd scenario, there were eight students aged between 22 and 24, with some occlusion present. Moreover, due to the camera's mounting position and field of view, the faces of students at very close range could not be identified. As shown in TABLE \ref{tab:result}, crowd recognition was difficult from a distance (output = free), but the presence of a crowd could be confirmed when captured at close range. In Case 3, due to the front-mounted camera, the crowd quickly moves out of the field of view, making its detection challenging.

\begin{figure}[htb]
    \centering
    \begin{minipage}[b]{0.4\linewidth}
        \centering
        \includegraphics[width=\linewidth]{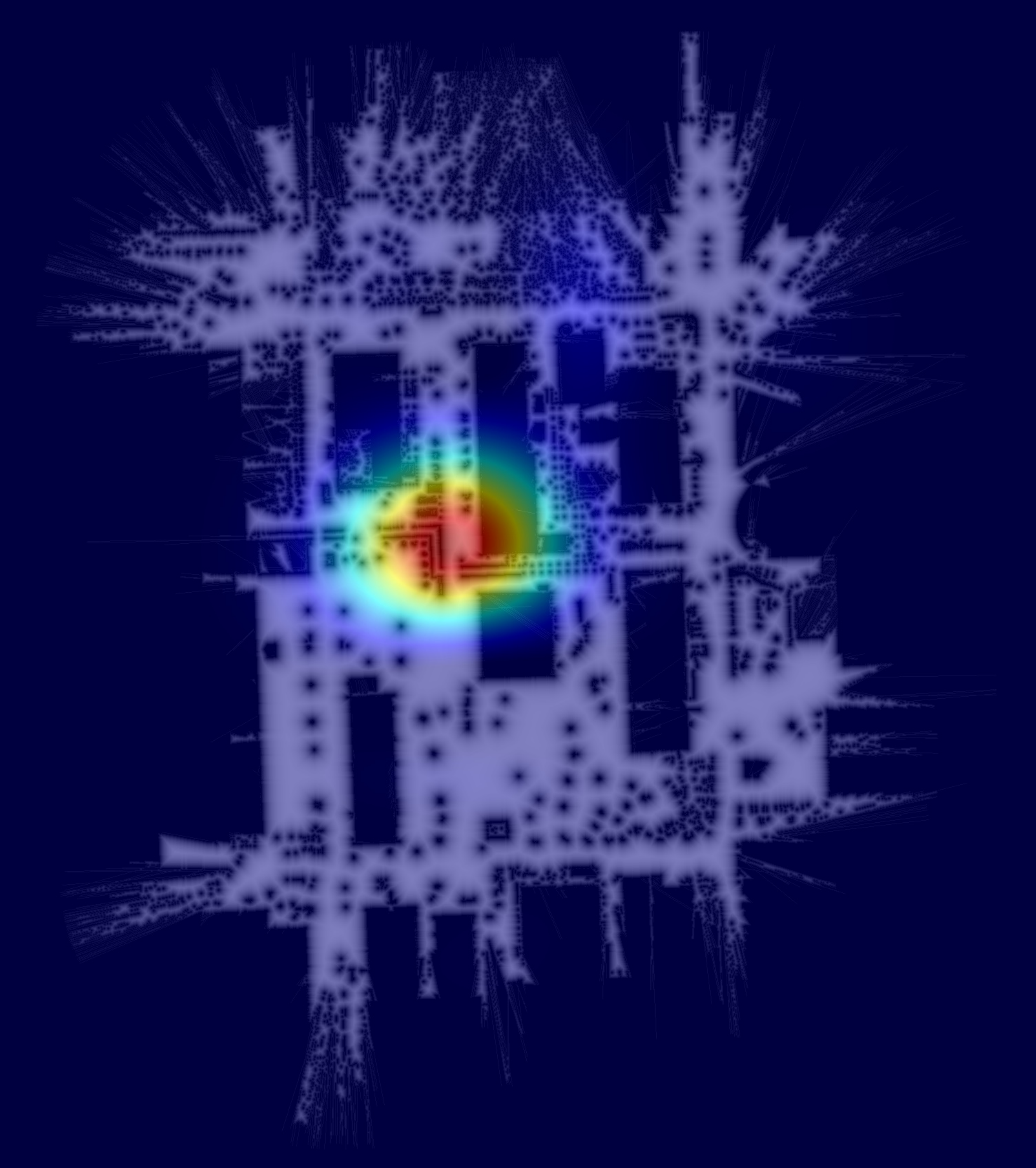}
        \subcaption{Trial 1}
        \label{fig:result_2a}
    \end{minipage}
    \begin{minipage}[b]{0.4\linewidth}
        \centering
        \includegraphics[width=\linewidth]{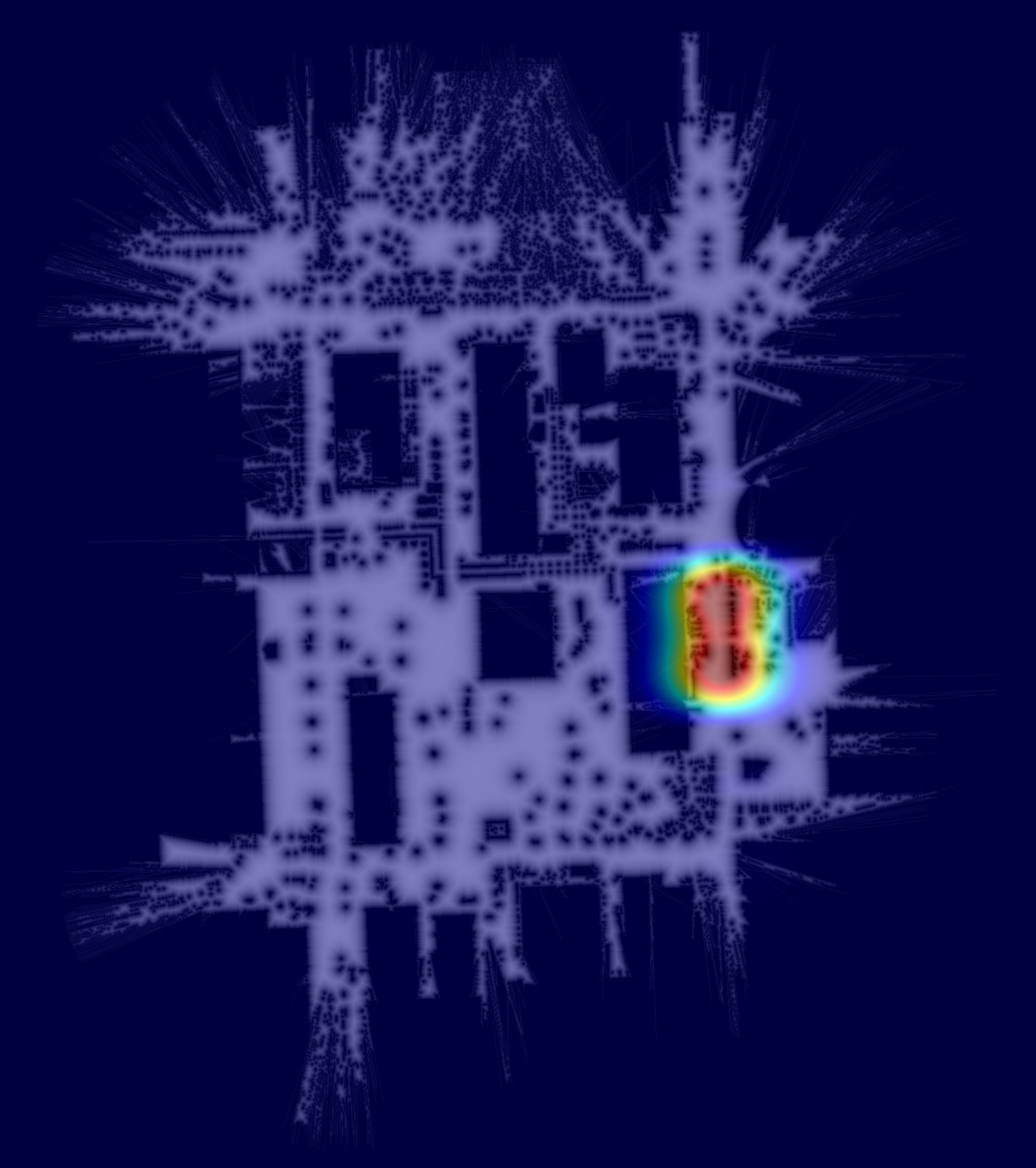}
        \subcaption{Trial 2}
        \label{fig:result_2b}
    \end{minipage}
    \caption{crowd recognition result by VLM}
    \label{fig:result_2}
\end{figure}

Next, using the cost map obtained by this system, a crowd map is generated by AMG. Fig. \ref{fig:result_2}\subref{fig:result_2a} shows the cost map created from result in the first trial, and Fig. \ref{fig:result_2}\subref{fig:result_2b} shows that of the second trial. This map is displayed by overlaying the generated crowd map onto the geometric map. When creating the maps, equal weighting (1:1) is applied to the geometric and crowd information; thus, in the path planning process, the costs associated with crowds and obstacles are considered equivalent. The crowd map is interpolated using Gaussian Process Regression, which reflects ambiguity in the map. We found that the peak in the cost map is located closer to the robot than the actual point where the crowd was generated. This is because of that the camera is forward-facing, the crowd appears to be located further ahead than the robot's current position, indicating a discrepancy between the self-localization results and the camera's viewpoint.

\begin{figure}[htbp]
    \centering
    \begin{minipage}[b]{0.4\linewidth}
        \centering
        \includegraphics[width=\linewidth]{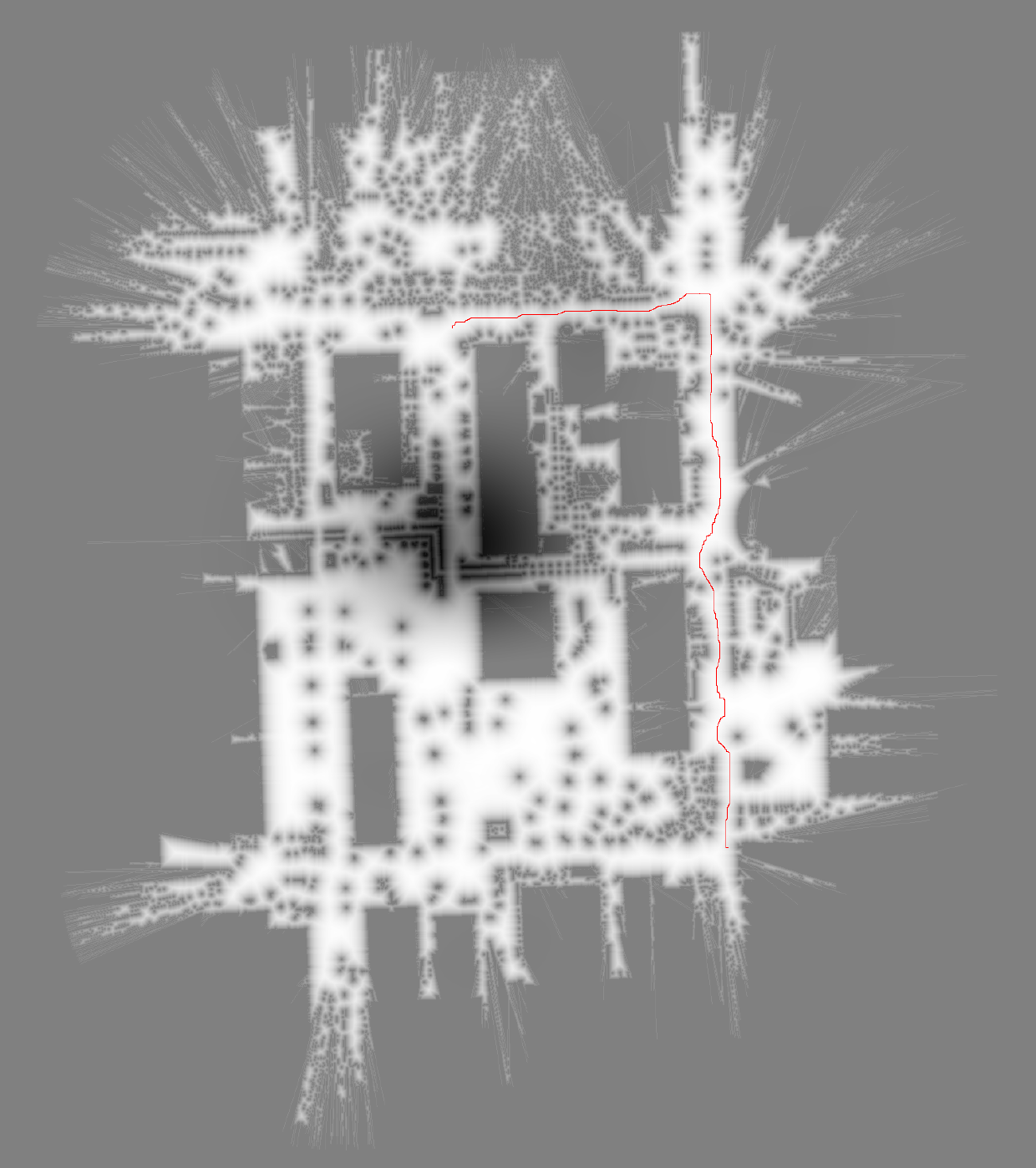}
        \subcaption{Trial 1}
        \label{fig:result_3a}
    \end{minipage}
    \begin{minipage}[b]{0.4\linewidth}
        \centering
        \includegraphics[width=\linewidth]{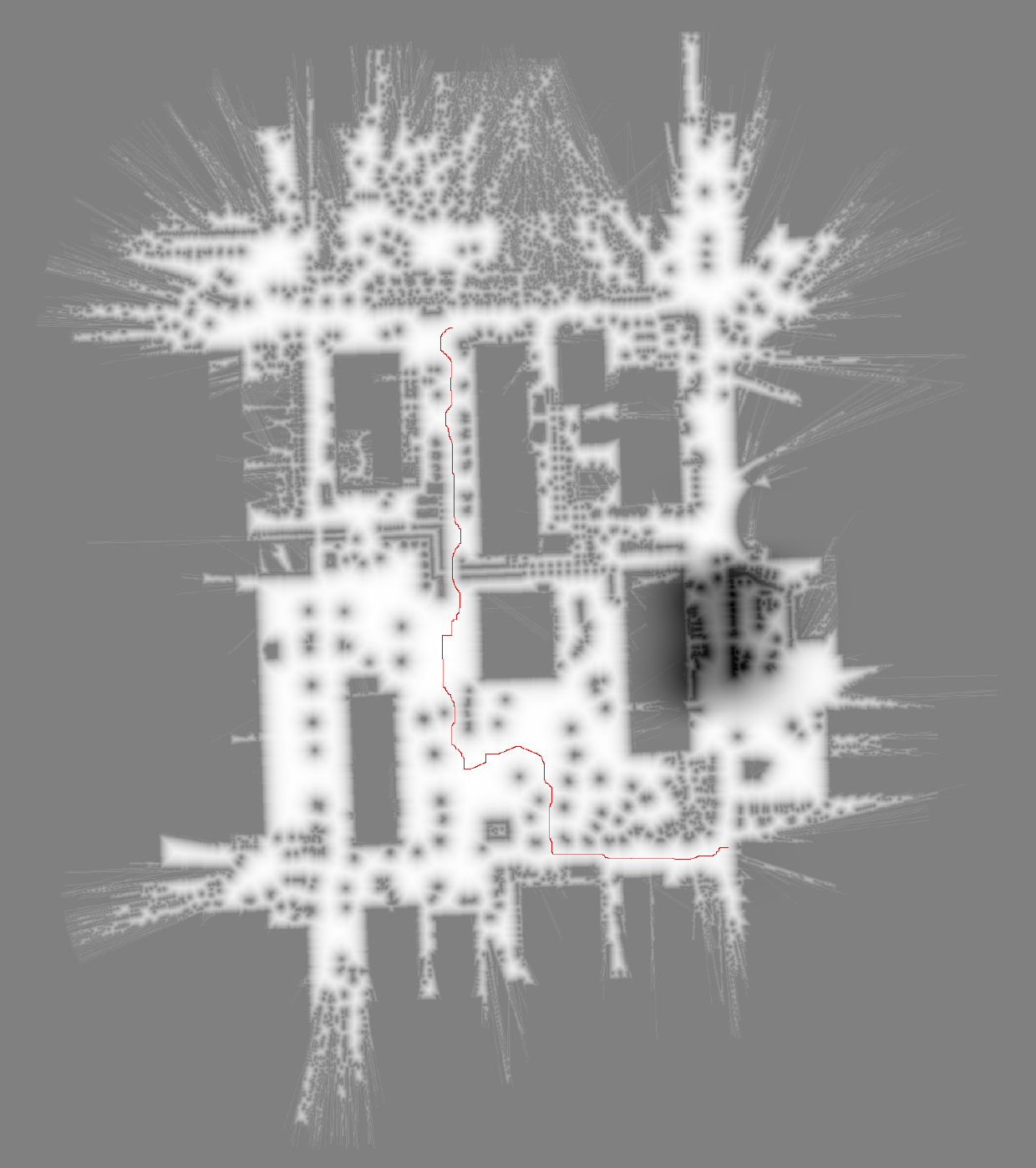}
        \subcaption{Trial 2}
        \label{fig:result_3b}
    \end{minipage}
    \begin{minipage}[b]{0.4\linewidth}
        \centering
        \includegraphics[width=\linewidth]{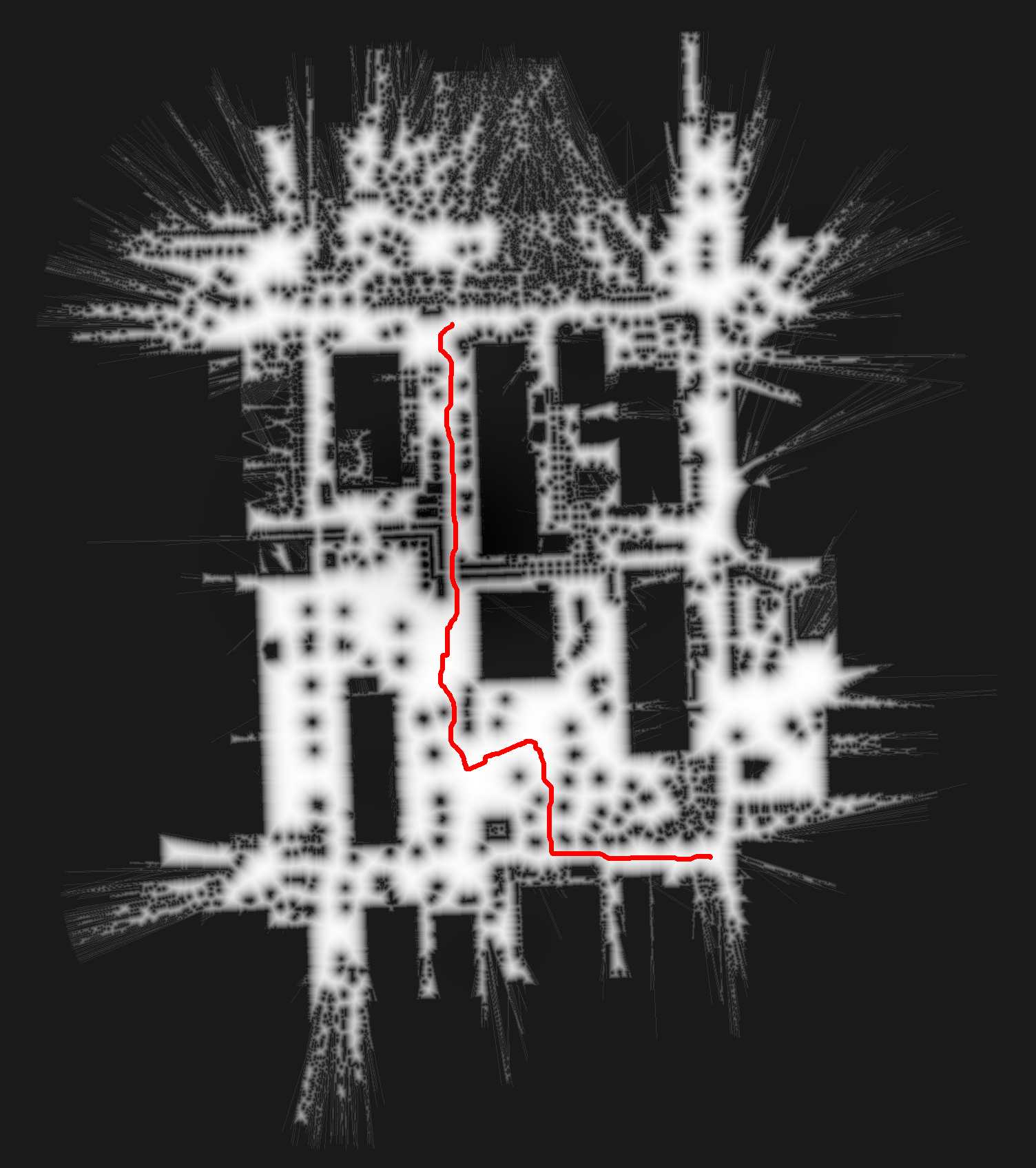}
        \subcaption{Trial 1 (geometoric:abstraction=9:1)}
        \label{fig:result_3c}
    \end{minipage}
    \caption{Generated path with Fig. \ref{fig:result_2}}
    \label{fig:result_3}
\end{figure}

Finally, using the crowd map shown in Fig. \ref{fig:result_2}, path planning is performed to reach the destination with avoiding crowds. Fig. \ref{fig:result_3}\subref{fig:result_3a} shows the result of path planning using Dijkstra's algorithm for the first trial, and Fig. \ref{fig:result_3}\subref{fig:result_3b} for the second trial, with the same start and end points. When generating the path, a grayscale map with integer values from 0 to 255 is used for path planning, which is lightweight and easy to implement. In both cases, it was confirmed that the robot successfully reached the destination while avoiding both crowds and static obstacles.
\textcolor{black}{The weighting of geometric information and crowd information requires careful adjustment. As an extreme example, in Fig. \ref{fig:result_3}\subref{fig:result_3c}, where the weighting of geometric information and crowd information is set at 1:9, we can see that a route is planned that neglects the cost of crowds.}
\textcolor{black}{In this experiment, we implemented navigation by considering the crowd as a nearly static environment. This suggests that it can be expanded to dynamic navigation by increasing the frequency of map and route updates. If the results of trial1 and trial2 are obtained consecutively as trials, it is considered that the maps of trial1 and trial2 will also require weighting similar to that of the geometric map. Figure o shows the results when the weight ratios are set to 1:1:1, 2:2:1, and 4:4:1, respectively. This weight balance should be appropriately adjusted according to the crowd flow model and the map update cycle.}

\begin{table*}[htb]
    \centering
    \caption{Crowd recognition result by VLM}
    \label{tab:result}
    \begin{tabular}{c|ccc}
        \hline
        \ & Case 1 & Case 2 & Case 3 \\
        \hline
        Image & 
        \begin{minipage}{5truecm}
            \centering
            \includegraphics[width=4.9truecm,clip]{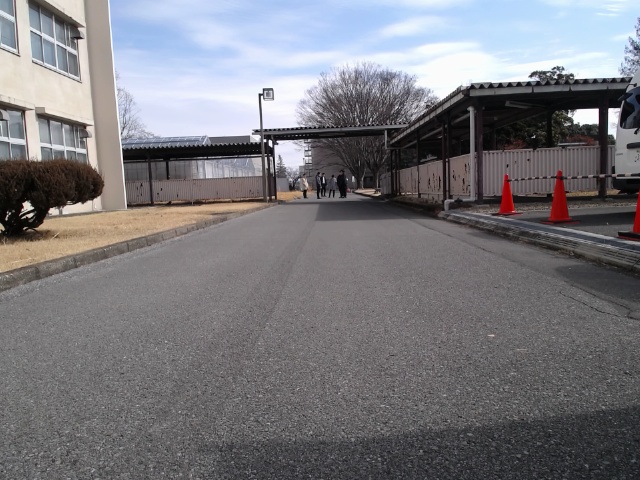}
        \end{minipage} &
        \begin{minipage}{5truecm}
            \centering
            \includegraphics[width=4.9truecm,clip]{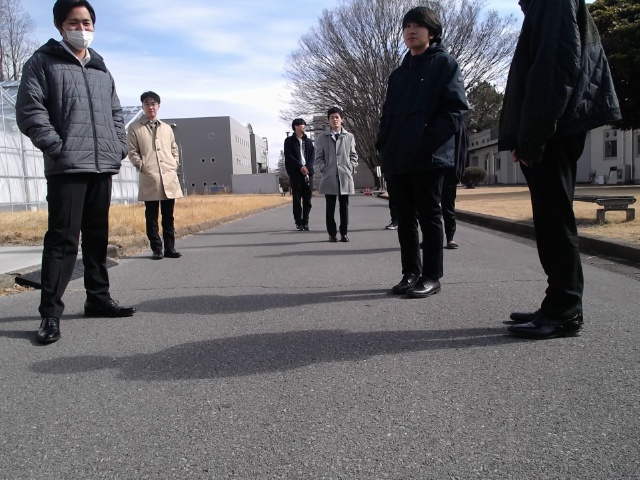}
        \end{minipage} &
        \begin{minipage}{5truecm}
            \centering
            \includegraphics[width=4.9truecm,clip]{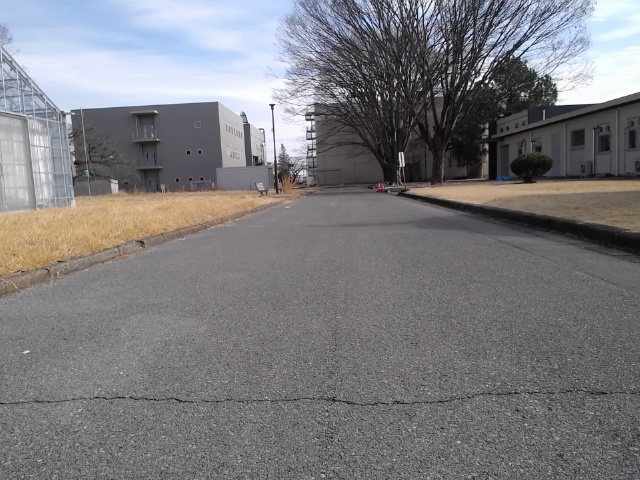}
        \end{minipage} \\
        Prompt & \multicolumn{3}{p{15truecm}}{You are an expert on robot drivability. The given image was taken by a camera placed in front of the robot. Analyze if there are any crowds in this image that would be an obstacle to the robot's traveling. If a crowd is determined to be present, return ``crowd''; otherwise, return ``free''.} \\
        \hline
        Output & free & crowd & free \\
        \hline
    \end{tabular}
\end{table*}

\begin{figure}[htbp]
    \centering
    \begin{minipage}[b]{0.4\linewidth}
        \centering
        \includegraphics[width=\linewidth]{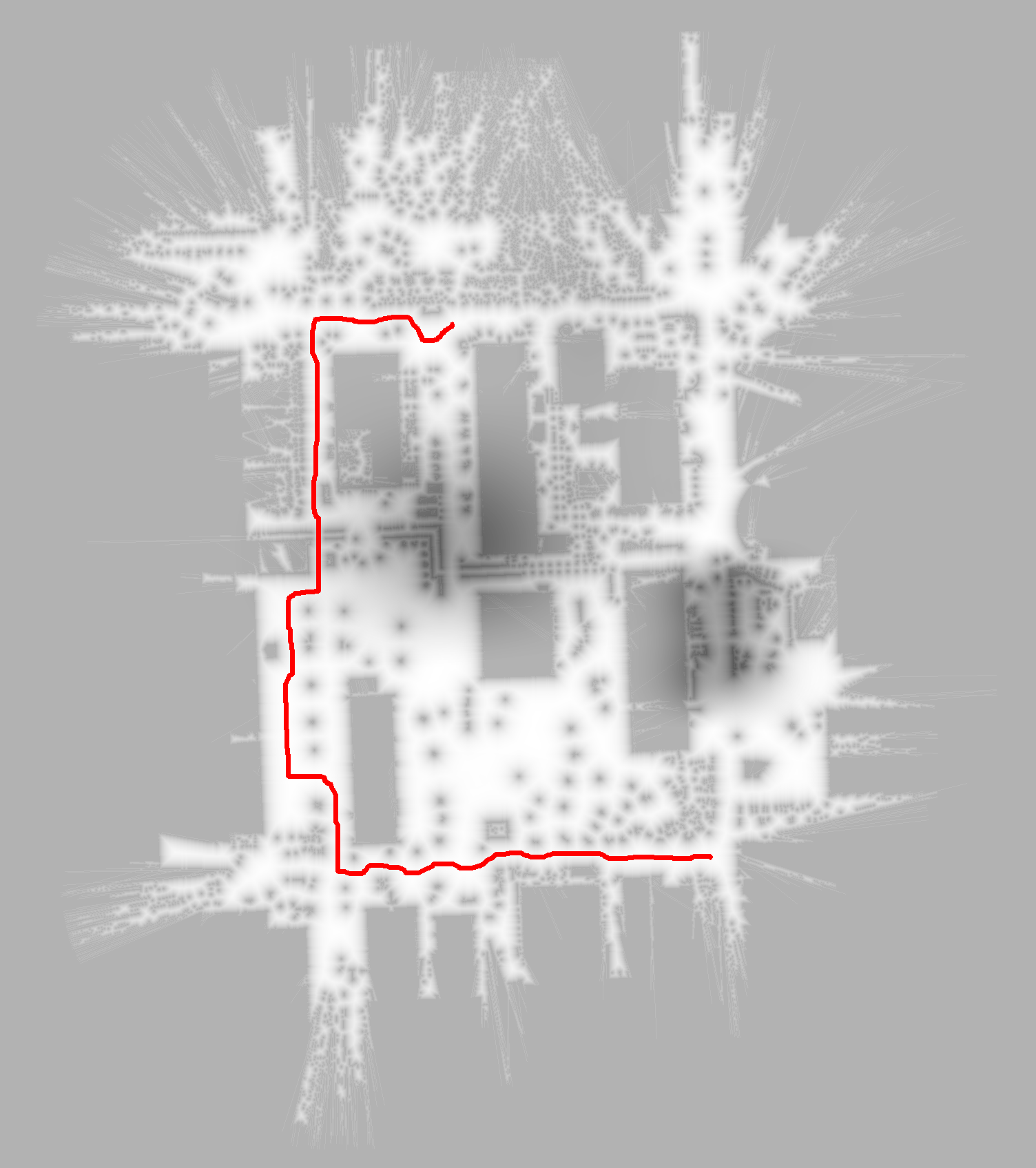}
        \subcaption{Geometoric:Trial1:Trial2=1:1:1}
        \label{fig:result_4a}
    \end{minipage}
    \begin{minipage}[b]{0.4\linewidth}
        \centering
        \includegraphics[width=\linewidth]{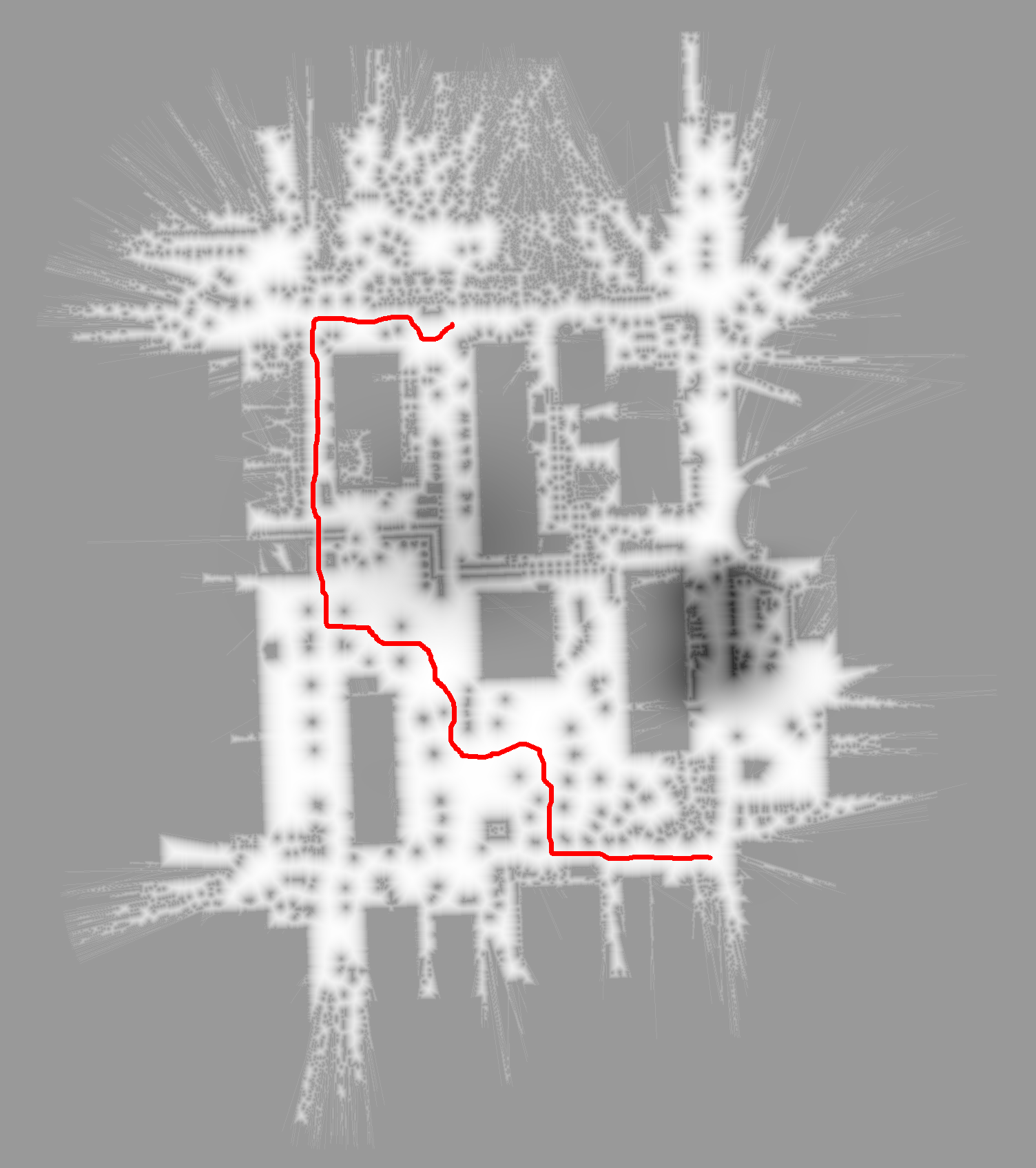}
        \subcaption{Geometoric:Trial1:Trial2=2:2:1}
        \label{fig:result_4b}
    \end{minipage}
    \begin{minipage}[b]{0.4\linewidth}
        \centering
        \includegraphics[width=\linewidth]{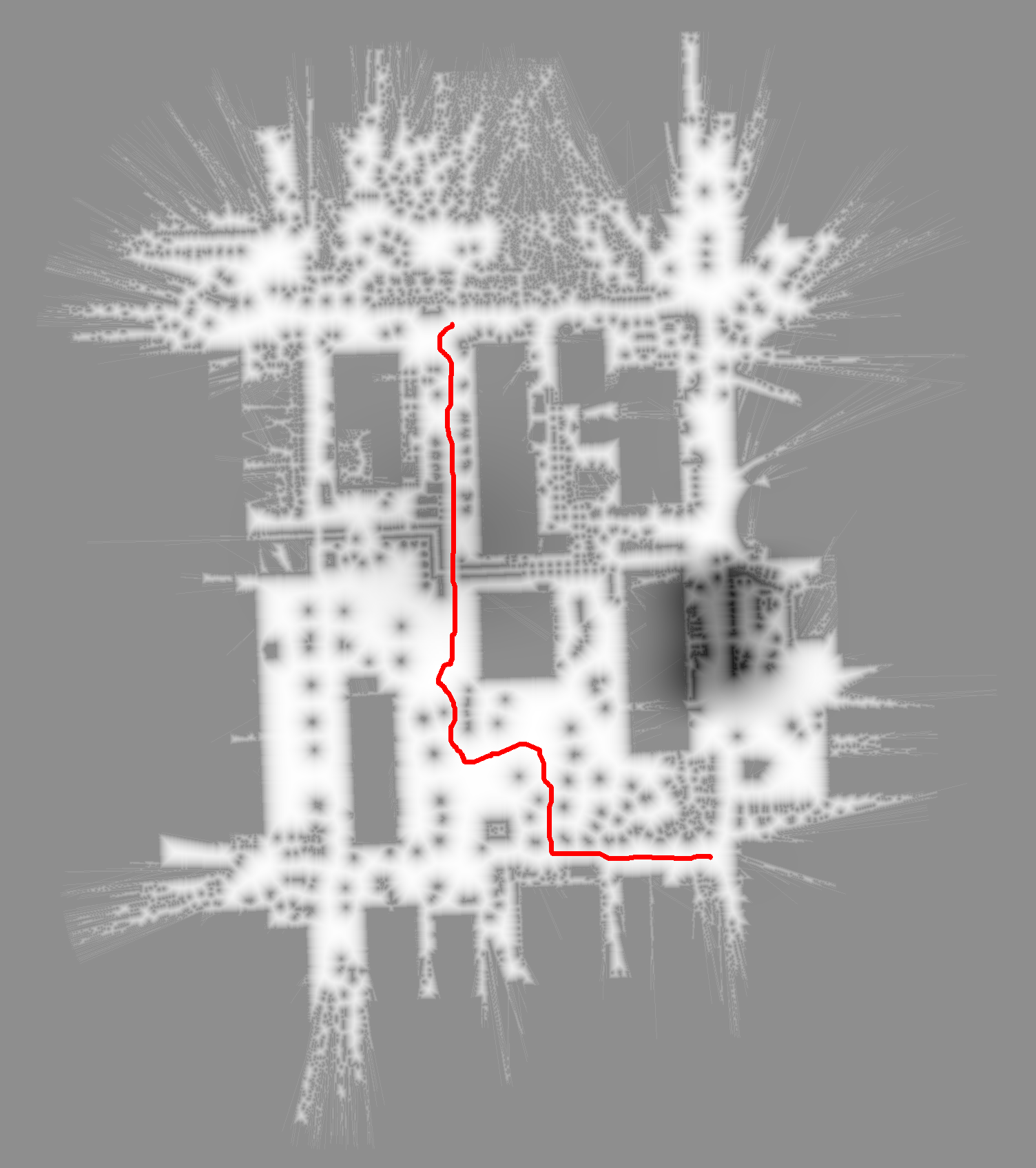}
        \subcaption{Geometoric:Trial1:Trial2=4:4:1}
        \label{fig:result_4c}
    \end{minipage}
    \caption{Paths generated by considering the execution of Fig. \ref{fig:result_2} as continuous time.}
    \label{fig:result_3}
\end{figure}

\begin{figure}[t]
    \centering
    \begin{minipage}{0.4\linewidth}
        \centering
        \includegraphics[width=\linewidth]{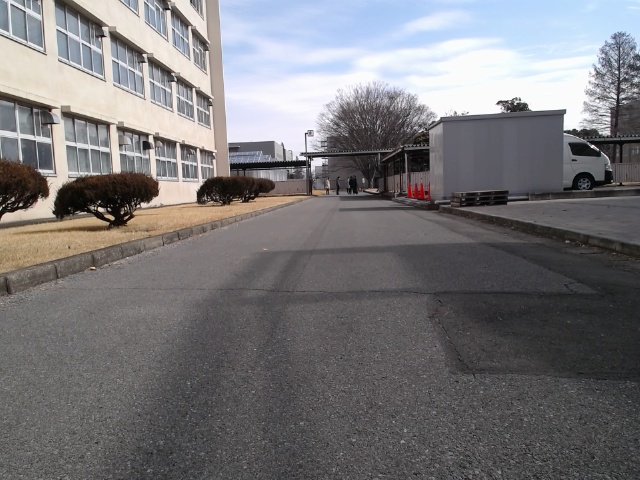}
        \subcaption{Far from the Crowd}
        \label{fig:discution_1}
    \end{minipage}
    \begin{minipage}{0.4\linewidth}
        \centering
        \includegraphics[width=\linewidth]{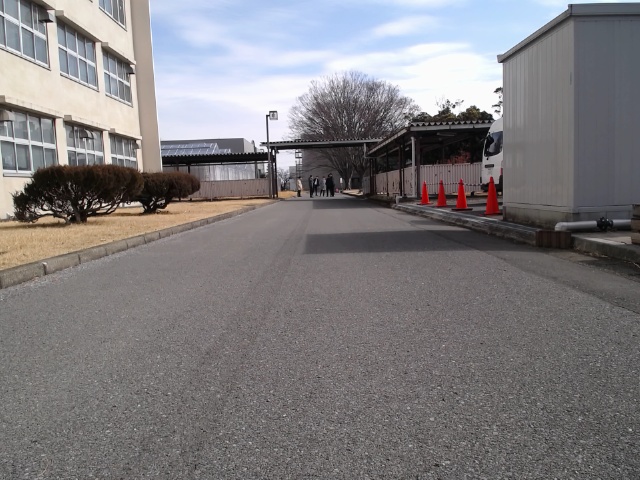}
        \subcaption{Near the Crowd}
        \label{fig:discution_2}
    \end{minipage}
    \begin{minipage}{0.4\linewidth}
        \centering
        \includegraphics[width=\linewidth]{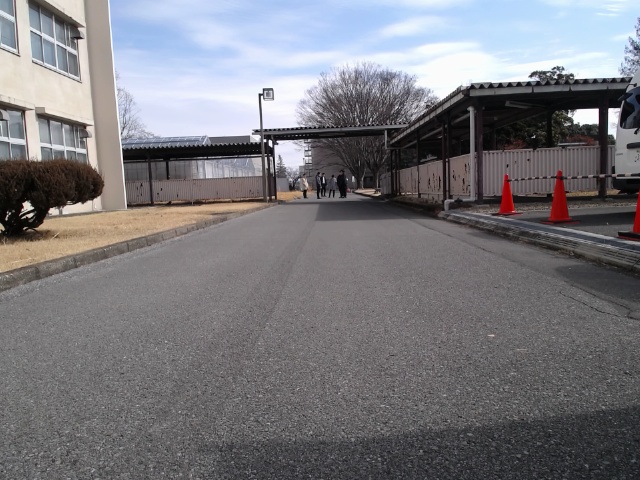}
        \subcaption{Closer to the Crowd}
        \label{fig:discution_3}
    \end{minipage}
    \caption{Unstable recognition pattern in case of crowd detection}
    \label{fig:frequntEvo}
\end{figure}

\section{Discussion}
Based on the experimental results, the effectiveness of the proposed method is discussed. First, when determining whether a particular location is classified as ``crowd'' or ``non-crowd'', there were some positions where the decision was unstable. Fig. \ref{fig:frequntEvo} shows examples of such images: Fig. \ref{fig:frequntEvo}\subref{fig:discution_1} is taken from the furthest point from the crowd, and Fig. \ref{fig:frequntEvo}\subref{fig:discution_3} from the closest. Although this may partly be influenced by the precision of the prompt, it is believed that the VLM being designed for abstract rather than precise spatial recognition produced these results. \textcolor{black}{Wen et al. show that GPT-4V has a scene understanding capability that exceeds that of conventional autonomous driving systems \cite{wen}. In nighttime scenarios, it was confirmed that the accuracy of distant object recognition decreases due to the limited visibility range, but by correctly understanding interference factors such as nighttime and long distances, GPT-4V can dynamically correct the reliability of the detection results and reduce the risk of falling into an incorrect local optimum solution.} Based on this data, the generated cost map is shown in Fig. \ref{fig:result_2a}. By utilizing the confidence derived from Gaussian Process Regression, it became possible to quantitatively evaluate, to some extent, the base model's level of abstract understanding. This suggests that combining the quantification of abstract concepts with the interpolation process of GPR is both effective and compatible. The Gaussian kernel is capable of representing the instability inherent in abstract concepts, this feature capable of quantification and mapping of another abstract information.

Regarding the path planning results using the crowd map, the outcomes were functioning. Although the 1:1 weighting was determined through trial and error, the generated path successfully avoided both crowds and obstacles while achieving the objective. This confirms that the proposed method functions effectively throughout the system.

\section{Conclusion}
In this study, we proposed a new method that utilizes Vision-Language Models (VLMs) to analyze abstract environmental information (crowds) and converts that information into a continuous cost map using Gaussian Process Regression (GPR). Conventional robot navigation methods have struggled to accurately analyze dynamic obstacles and abstract concepts such as crowds and to effectively integrate these into path planning. However, our proposed method leverages the integrated recognition capabilities of VLMs to generate an abstract environmental map that functions well even in real-world settings. Experiments conducted on the Utsunomiya University campus confirmed that the path planning based on our method can effectively avoid crowds, resulting in safer and more efficient navigation.

Future work includes generating multi-layer maps that integrate various abstract environmental factors beyond crowds—such as road conditions, lighting, noise levels, and hazardous areas—and developing comprehensive path planning algorithms that utilize these maps. Achieving such multi-layer integration will require appropriately modeling the correlations between different environmental factors and developing methods for efficient, real-time integration. Furthermore, further investigation into the accuracy and stability of abstract information analysis using VLMs is needed.

\addtolength{\textheight}{-12cm}




\printbibliography

@INPROCEEDINGS{depatla,
  author={Depatla, Saandeep and Mostofi, Yasamin},
  booktitle={2018 IEEE International Conference on Pervasive Computing and Communications (PerCom)}, 
  title={Crowd Counting Through Walls Using WiFi}, 
  year={2018},
  volume={},
  pages={1-10}
}

@article{gong,
title = "Counting people in the crowd using social media images for crowd management in city events",
author = "X. Gong and W. Daamen and Alessandro Bozzon and S.P. Hoogendoorn",
year = "2021",
volume = "48",
pages = "3085--3119",
journal = "Transportation"
}

@ARTICLE{choi,
  author={Choi, Hyuckjin and Fujimoto, Manato and Matsui, Tomokazu and Misaki, Shinya and Yasumoto, Keiichi},
  journal={IEEE Access}, 
  title={Wi-CaL: WiFi Sensing and Machine Learning Based Device-Free Crowd Counting and Localization}, 
  year={2022},
  volume={10}
}

@article{jiang,
title = {Indoor occupancy estimation from carbon dioxide concentration},
journal = {Energy and Buildings},
volume = {131},
pages = {132-141},
year = {2016},
author = {Chaoyang Jiang and Mustafa K. Masood and Yeng Chai Soh and Hua Li}
}

@Article{dijkstra,
author={Dijkstra, E. W.},
title={A note on two problems in connexion with graphs},
journal={Numerische Mathematik},
year={1959},
volume={1},
pages={269-271}
}

@INPROCEEDINGS{trautman1,
  author={Trautman, Peter and Krause, Andreas},
  booktitle={2010 IEEE/RSJ International Conference on Intelligent Robots and Systems}, 
  title={Unfreezing the robot: Navigation in dense, interacting crowds}, 
  year={2010},
  volume={},
  pages={797-803}
}

@INPROCEEDINGS{trautman2,
  author={Trautman, Peter and Ma, Jeremy and Murray, Richard M. and Krause, Andreas},
  booktitle={2013 IEEE International Conference on Robotics and Automation}, 
  title={Robot navigation in dense human crowds: the case for cooperation}, 
  year={2013},
  volume={},
  pages={2153-2160}
}

@INPROCEEDINGS{changan,
  author={Chen, Changan and Liu, Yuejiang and Kreiss, Sven and Alahi, Alexandre},
  booktitle={2019 International Conference on Robotics and Automation (ICRA)}, 
  title={Crowd-Robot Interaction: Crowd-Aware Robot Navigation With Attention-Based Deep Reinforcement Learning}, 
  year={2019},
  volume={},
  pages={6015-6022}
}

@article{anirudh,
  author       = {Anirudh Vemula and
                  Katharina M{\"{u}}lling and
                  Jean Oh},
  title        = {Modeling Cooperative Navigation in Dense Human Crowds},
  journal      = {CoRR},
  volume       = {abs/1705.06201},
  year         = {2017}
}

@INPROCEEDINGS{liu,
  author={Liu, Lucia and Dugas, Daniel and Cesari, Gianluca and Siegwart, Roland and Dubé, Renaud},
  booktitle={2020 IEEE/RSJ International Conference on Intelligent Robots and Systems (IROS)}, 
  title={Robot Navigation in Crowded Environments Using Deep Reinforcement Learning}, 
  year={2020},
  volume={},
  pages={5671-5677}
}

@INPROCEEDINGS{henry,
  author={Henry, Peter and Vollmer, Christian and Ferris, Brian and Fox, Dieter},
  booktitle={2010 IEEE International Conference on Robotics and Automation}, 
  title={Learning to navigate through crowded environments}, 
  year={2010},
  volume={},
  pages={981-986}
}

@ARTICLE{truong,
  author={Truong, Xuan-Tung and Ngo, Trung Dung},
  journal={IEEE Transactions on Automation Science and Engineering}, 
  title={Toward Socially Aware Robot Navigation in Dynamic and Crowded Environments: A Proactive Social Motion Model}, 
  year={2017},
  volume={14},
  pages={1743-1760}
}

@ARTICLE{shami,
  author={Shami, Mamoona Birkhez and Maqbool, Salman and Sajid, Hasan and Ayaz, Yasar and Cheung, Sen-Ching Samson},
  journal={IEEE Transactions on Circuits and Systems for Video Technology}, 
  title={People Counting in Dense Crowd Images Using Sparse Head Detections}, 
  year={2019},
  volume={29},
  pages={2627-2636}
}

@misc{ma,
      title={Bayesian Loss for Crowd Count Estimation with Point Supervision}, 
      author={Zhiheng Ma and Xing Wei and Xiaopeng Hong and Yihong Gong},
      year={2019}
}

@article{sharma,
title = {Scale-aware CNN for crowd density estimation and crowd behavior analysis},
journal = {Computers and Electrical Engineering},
volume = {106},
pages = {108569},
year = {2023},
author = {Vipal Kumar Sharma and Roohie Naaz Mir and Chandrapal Singh}
}

@article{seeger,
author = {Seeger, Matthias},
title = {Gaussian processes for machine leearning},
journal = {International Journal of Neural Systems},
volume = {14},
pages = {69-106},
year = {2004}
}

@article{akaho,
  title={Introduction to Gaussian Process Regression},
  author={Shotaro Akaho},
  journal={SYSTEMS, CONTROL AND INFORMATION},
  volume={62},
  pages={390-395},
  year={2018}
}

@ARTICLE{alatise,
  author={Alatise, Mary B. and Hancke, Gerhard P.},
  journal={IEEE Access}, 
  title={A Review on Challenges of Autonomous Mobile Robot and Sensor Fusion Methods}, 
  year={2020},
  volume={8},
  pages={39830-39846}
}

@misc{wen,
      title={On the Road with GPT-4V(ision): Early Explorations of Visual-Language Model on Autonomous Driving}, 
      author={Licheng Wen and Xuemeng Yang and Daocheng Fu and Xiaofeng Wang and Pinlong Cai and Xin Li and Tao Ma and Yingxuan Li and Linran Xu and Dengke Shang and Zheng Zhu and Shaoyan Sun and Yeqi Bai and Xinyu Cai and Min Dou and Shuanglu Hu and Botian Shi and Yu Qiao},
      year={2023},
}

\end{document}